\documentclass{article}
\usepackage{ijcai20}

\usepackage{times}
\usepackage{soul}
\usepackage{url}
\usepackage[hidelinks]{hyperref}
\usepackage[utf8]{inputenc}
\usepackage[small]{caption}
\usepackage{graphicx}
\usepackage{amsmath}
\usepackage{amsthm}
\usepackage{booktabs}
\usepackage{algorithm}
\usepackage{algorithmic}
\usepackage{amsfonts}
\title{Co-Saliency Spatio-Temporal Interaction Network for Person  Re-Identification \\ in Videos}

\author{
	Jiawei Liu$^{1}$\and
	Zheng-Jun Zha$^{1*}$\and
	Xierong Zhu$^{1}$\And
	Na Jiang$^2$\\
	\affiliations
	$^1$University of Science and Technology of China, China\\
	$^2$Capital Normal University, China\\
	\emails
	\{jwliu6,zhazj\}@ustc.edu.cn, zxr8192@mail.ustc.edu.cn, jiangna@cnu.edu.cn
}

\begin{document}

\maketitle

\begin{abstract}
 Person re-identification aims at identifying a certain pedestrian across non-overlapping camera networks. Video-based person re-identification approaches have gained significant attention recently, expanding image-based approaches by learning features from multiple frames. In this work, we propose a novel Co-Saliency Spatio-Temporal Interaction Network (CSTNet) for person re-identification in videos. It captures the common salient foreground regions among video frames and explores the spatial-temporal long-range context interdependency from such regions, towards learning discriminative pedestrian representation. Specifically, multiple co-saliency learning modules within CSTNet are designed to utilize the correlated information across video frames to extract the salient features from the task-relevant regions and suppress background interference. Moreover, multiple spatial-temporal interaction modules within CSTNet are proposed, which exploit the spatial and temporal long-range context interdependencies on such features and spatial-temporal information correlation, to enhance feature representation. Extensive experiments on two benchmarks have demonstrated the effectiveness of the proposed method.
\end{abstract}

\section{Introduction}

\renewcommand{\thefootnote}{}
\footnotetext{* Corresponding author}
Person re-identification aims to search for a person-of-interest across non-overlapping camera networks. It has drawn significant attention recently owing to its importance in many practice applications, such as automated surveillance, activity analysis and content-based visual retrieval \textit{etc} \cite{hou2019vrstc,liu2016multi,zhang2014robust}. Despite recent progress in person re-identification, it remains a very challenging task due to background clutter, occlusion, as well as dramatic variations in human poses, illuminations and camera viewpoints \textit{etc} \cite{liu2019adaptive,liu2019dense,zhang2012attribute}.

\begin{figure}[!t]
	\begin{center}
		\includegraphics[width=0.9\linewidth]{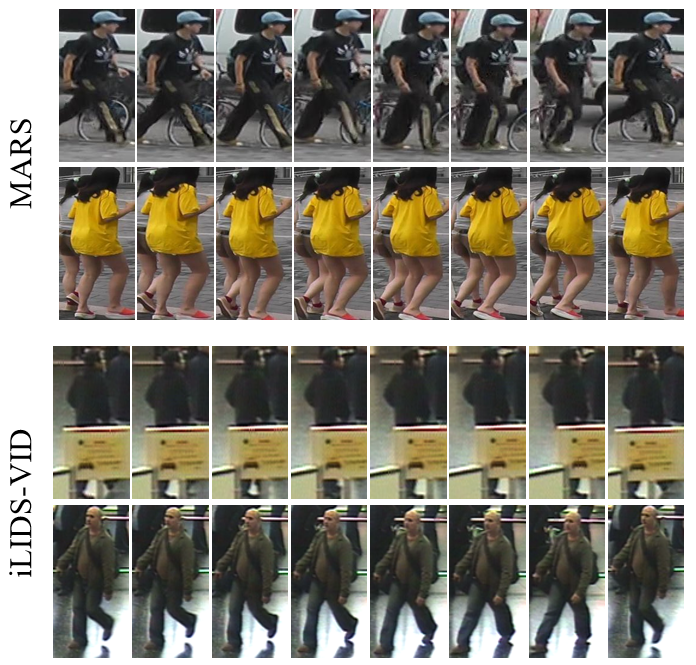}
	\end{center}
	\caption{Example video sequences in the MARS and iLIDS-VID video person re-identification datasets.}
	\label{fig1}
\end{figure}

Generally, person re-identification is approached with either image or video data of pedestrian for representation. Recent years have witnessed the impressive progresses in image-based person re-identification, \textit{e.g.}, deep representations have significantly boosted the re-identification performance on the image datasets. However, image-based approaches are easily susceptible to the quality of pedestrian images, since limited amount of information of a single image often results in a lower tolerance to the visual ambiguity in intra-class and inter-class appearance. In contrast, a video sequence contains richer spatial and temporal cues of pedestrians, which are important for identifying pedestrians. Thus, video-based person re-identification has better potentials to address such challenges in image-based person re-identification. Figure~1 illustrates some sample video sequences of pedestrians in two real-world datasets, \textit{i.e.}, MARS \cite{zheng2016mars} and iLIDS-VID \cite{wang2014person}.  


For re-identifying pedestrians in videos, one key issue is to extract discriminative video-level visual representation. Many of video-based person re-identification methods extract frame-level appearance features by considering the whole video frames followed by temporal feature aggregation, \textit{e.g.}, pooling operation/RNN \cite{mclaughlin2016recurrent,hermans2017defense}. Unfortunately, these methods often fail due to severe occlusions and background clutter in the video datasets. It is a high probability that noisy background information from irrelevant non-salient regions of video frames may get misinterpreted as useful appearance features, leading to a drastic drop in re-identification accuracy. Hence, some works \cite{zhao2017spindle,li2018harmonious} exploit augmented information such as human pose estimation to focus on the subject pedestrian and avoid extracting visual cues from  background for generating more effective representations. However, surveillance video sequences often exist drastic viewpoint and body pose variations as well as low-resolution, resulting in unstable human pose estimation. Moreover, these methods only locate pedestrian’s key joint locations, and miss out the accessories associated with pedestrians (\textit{e.g.} hat, backpack, bike), which are also significant cues for matching. Instead of using such expensive augmented information, some other works \cite{li2018diversity,zhou2017see,xu2017jointly} utilize attention mechanism to concentrate on the subject pedestrian, which discover distinctive body parts by using diverse spatial attentions, and crucial frames by using temporal attention. Nevertheless, these attention-based approaches are also sub-optimal since their models work on “per-frame” basis, thus under-utilizing the rich spatial-temporal information available in videos. 

On the other hand, many empirical studies \cite{wang2018non} have suggested that the performance can be greatly improved by proper modeling of context interdependency for feature maps in recognition tasks. Nevertheless, current video-based methods do not make best of spatial-temporal context interdependency modeling. The typical 2D convolution operation completely neglect the temporal information of pedestrian videos. The 3D convolution operation and its variations can capture spatail-temporal context interdependency, but they are limited to local context interdependency modeling. A few methods \cite{rao2019non,liao2018video} directly employ non-local operation \cite{wang2018non} on video data to capture spatial-temporal long-range context interdependency, but they suffer from huge computation complexity and could not make full use of the correlation between spatial and temporal information.

\begin{figure*}[!t]
	\begin{center}
		\includegraphics[width=0.92\linewidth]{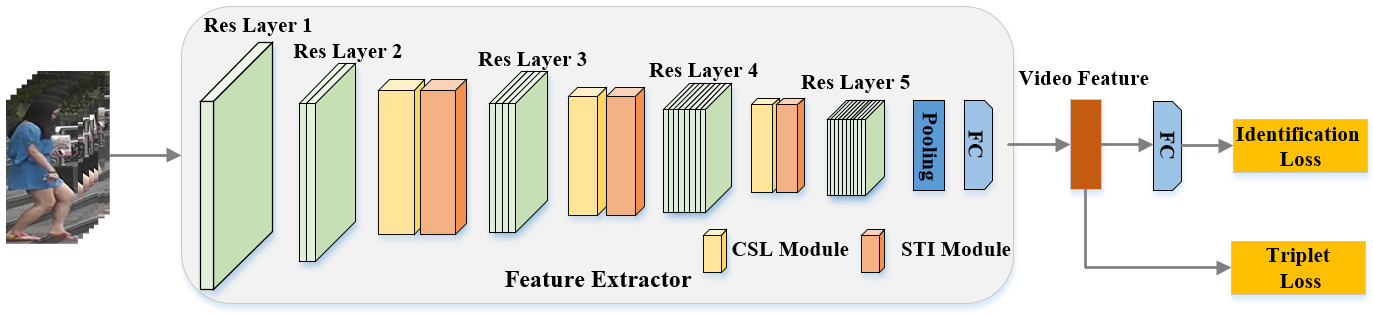}
	\end{center}
	\caption{The overall architecture of the proposed CSTNet. It consists of multiple co-saliency learning modules, multiple spatial-temporal interaction modules as well as a backbone network.}
	\label{img2}
\end{figure*}

In this work, we propose a novel Co-Saliency Spatio-Temporal Interaction Network (CSTNet) for person re-identification in videos. It captures the common salient foreground regions among video frames and models the spatial-temporal long-range context interdependency for the feature maps of such salient regions, to learn discriminative pedestrian representation. As illustrated in Figure~\ref{img2}, CSTNet consists of multiple co-saliency learning (CSL) modules, multiple spatial-temporal interaction (STI) modules, as well as a backbone network containing several residual layers \cite{he2016deep}. Each co-saliency learning module is followed with a spatial-temporal interaction module, and they are plugged between different residual layers. The co-saliency learning module utilizes normalized cross correlation algorithm \cite{subramaniam2016deep} to learn the spatial-channel attention map corresponding to the foreground pedestrian (along with its accessories) for each video frame by consulting with all video frames. It then extracts the common salient features from video frames and suppresses irrelevant background information. The spatial-temporal interaction module discovers two types of long-range interdependencies: the spatial relation which models interdependency between positions in the salient feature map of a single frame,  and the temporal relation which models the interdependency between the same position across all frames, to generate spatial and temporal relation features. A fusion operation is designed to learn the correlation between spatial and temporal relation features and aggregate them into  discriminative pedestrian representation. We conduct extensive experiments to evaluate the proposed CSTNet on two challenging benchmarks, \textit{i.e.}, MARS and iLIDS-VID, and report superior performance over state-of-the-art approaches.

The main contribution of this paper is three-fold: (1) We propose a novel Co-Saliency Spatio-Temporal Interaction Network for person re-identification in videos. (2) We develop multiple co-saliency learning modules for capturing the common salient regions among video frames and suppressing background interference. (3) We design multiple spatial-temporal interaction modules to explore the spatial-temporal long-range context interdependency from such regions.

\section{Related Work}




\paragraph{Image-based Person Re-identification.} Conventional approaches for image-based person re-identification mainly focus on designing hand-crafted descriptors and/or learning appropriate distance metric. Recently, deep learning technique has been applied for person re-identification, towards learning discriminative pedestrian representation. For example,  Zhao \textit{et al.} \cite{zhao2017spindle} proposed the Spindle Net, which captures semantic features from different body regions with a region proposal network, and learns aligned features to address pose variations. Li \textit{et al.}  \cite{li2018harmonious} formulated a Harmonious Attention CNN (HA-CNN) for the joint learning of soft pixel attention and hard region attention.

\paragraph{Video-based Person Re-identification.} Early works on video-based person re-identification focus on hand-crafted video representations and/or distance metric learning. Recent approaches are mostly based on deep learning techniques. Some methods are developed as a straightforward extension of image-based re-identification method. They extracted pedestrian feature from each frame by various CNN models, and aggregated frame-level features across time by pooling operation or RNNs \cite{mclaughlin2016recurrent,hermans2017defense}. For learning more effective representations from foreground regions of video frames, attention mechanism and human pose estimation algorithm have been applied to lots of person re-identification methods \cite{xu2017jointly,song2018mask}. For example, Li \textit{et al.} \cite{li2018diversity} formulated a spatial-temporal attention model, which learns multiple spatial attention models and employs a diversity regularization term to ensure multiple models do not discover the same body part. Moreover, a few works attempt to utilize the long-range context interdependency by non-local operation for enhancing feature representation. For example, Liao \textit{et al.} \cite{liao2018video} proposed an end-to-end 3D ConvNet with non-local architecture, which integrates a spatial-temporal attention to aggregate a discriminative representation from a video sequence of pedestrian.  

\section{Method}

In this section, we firstly present the overall architecture of the proposed CSTNet, and then introduce each component of  CSTNet in the following subsections.

\subsection{Overall Architecture}


Given a training set $\boldsymbol{X} = \{\boldsymbol{x}_i\}_{i=1}^N$  containing $N$ video sequences from $K$ pedestrians captured by non-overlapping camera networks together with their corresponding person ID as $\boldsymbol{Y} = \{\boldsymbol{y}_i\}_{i=1}^N$, the objective is to learn discriminative representations from video sequences for identifying the same pedestrian and distinguishing different pedestrians. We propose a novel Co-Saliency Spatio-Temporal Interaction Network for person re-identification in videos, which captures the common salient foreground regions among video frames and explores the spatial-temporal long-range context interdependency from such salient regions to learn discriminative pedestrian representation. As shown in Figure~\ref{img2}, CSTNet consists of a backbone network, three co-salience learning modules, three spatial-temporal interaction modules and a classifier. The proposed CSTNet samples $T$ frames from all frames of a video sequence to form a video clip as input, which is then fed into the architecture for extracting pedestrian representation. Specifically, the backbone network contains five residual layers, which is built on ResNet-50 model\cite{he2016deep} due to its strong ability in learning visual representation. The co-saliency learning modules are plugged after the second, third and fourth residual layers, which are designed for extracting the salient feature maps from video frames and suppressing irrelevant background information. Each co-saliency learning module is followed with a spatial-temporal interaction module, which is developed to exploit the spatial-temporal long-range context interdependency on the salient feature maps for enhancing the capacity of feature representation. At the end of the five residual layers, the average pooling layer is applied to summarize the frame-level descriptors to a clip-level feature. The resulting clip-level feature is taken into a classifier which is implemented with two fully connected layers to predict the probability that a video belongs to a particular person identity.

\subsection{Co-Saliency Learning Module}

\begin{figure}[!t]
	\begin{center}
		\includegraphics[width=1.0 \linewidth]{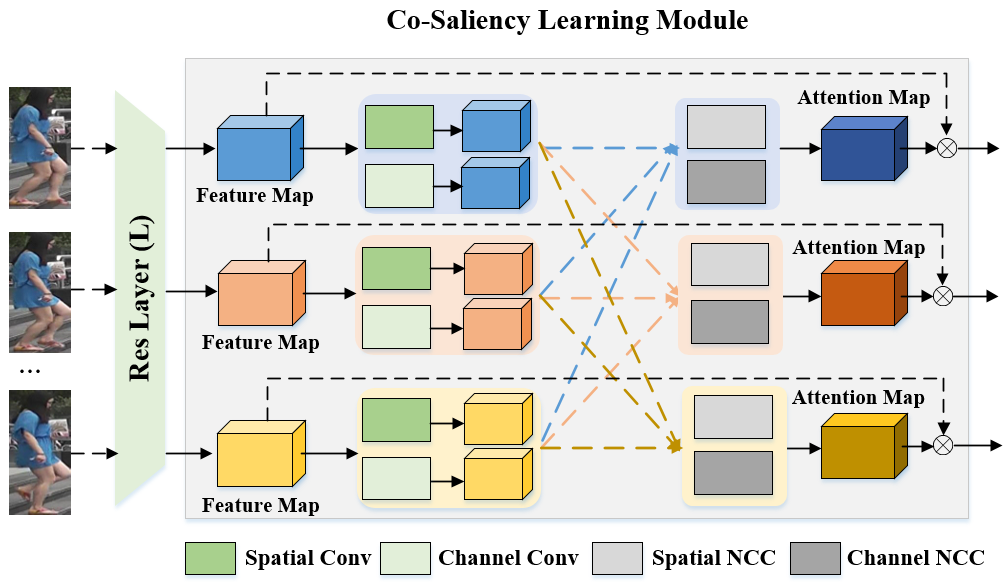}
	\end{center}
	\caption{Detailed structure of the co-saliency learning module.}
	\label{img3}
\end{figure}

The co-saliency learning module is designed to exploit the co-saliency inspired attention mechanism for extracting the salient features from the common foreground regions (pedestrian with its accessories) of video frames and suppressing irrelevant background information. The detailed architecture of the co-saliency learning module is shown in Figure~\ref{img3}. The input of this module is the set of frame-level feature maps of a pedestrian after a residual layer. The feature map is denoted by $\boldsymbol{f}_{t,y} \in \mathbb{R}^{C\times H\times W} $, where $t$ is the index of the video frame of the person identified by the label $y$, as well as  $C$, $H$ and $W$ denote the number of channels, height and width of the feature map, respectively.

The input feature maps $\{\boldsymbol{f}_{t,y}\}_{t=1}^T$ are firstly passed through a dimension reduction layer (implemented by two convolution layers followed with BN and ReLU) to obtain two types of feature maps with reduced dimension $C_L \times H\times W$ and $C \times H_L\times W_L$ ($ C_L \ll C$ and $ H_L\times W_L \ll H\times W$),  which greatly speed-up the computations. These two types of feature maps are then fed into the normalized cross correlation (NCC) blocks, which are used to estimate the spatial-channel attention map ($C \times H\times W$) for each frame by consulting with all video frames. The spatial-channel attention map activates the spatial locations of the foreground pedestrian of video frames and give more importance to the common important channels. For simplicity, we only introduce the detailed process of estimating the spatial attention map for each frame. Given the feature maps with dimension $C_L \times H\times W$, we consider the channel-wise feature vector at each spatial location  $(i, j) (1\leq i \leq H, 1\leq j \leq W)$ as a $C_L$ dimensional local descriptor of the frame at location $(i, j)$, denoted by $\boldsymbol{f}_{t,y}^{i,j}$. To match the local regions among video frames, for each spatial position $(i, j)$ of the frame, we compare the local descriptor $\boldsymbol{f}_{t,y}^{i,j}$ to all the local descriptors of other $(T-1)$ video frames. The comparison between the local descriptors is implemented by normalized cross correlation operation \cite{subramaniam2016deep}, which is robust to illumination variation. After that, the obtained comparison results are reshaped into a 3D volume with dimension ($(T-1)HW\times H\times W$),  where the value of each spatial position $(i, j)$ represents the correlation score. The 3D volume is defined as follows:
\begin{equation}
\begin{split}
Volume_{t,y}^{i,j} &=  NCC(\boldsymbol{f}_{t,y}^{i,j}, \boldsymbol{f}_{k,y}^{h,w}),  \\
&1\leq k \leq T,  k \ne t \\
&1\leq h \leq H \\
&1\leq w \leq W 
\end{split}
\end{equation}
The NCC operation is formulated as follows:
\begin{equation}
\begin{split}
NCC(P,Q) = \frac{1}{C_L} \frac{\sum_{g=1}^{C_L}(P_g-\mu_P)(Q_g-\mu_Q)}{\sigma_P\cdot \sigma_Q}   
\end{split}
\end{equation}
where $P$, $Q$ are two local descriptors, $(\mu_P, \mu_Q)$ and $(\sigma_P, \sigma_Q)$ denote the mean and standard deviation of the two descriptors, respectively. The obtained 3D volume is then summarized by using a convolution layer, and generates a spatial attention map $Z_s \in \mathbb{R}^ {1\times H\times W}$ for the video frame. Analogously, we can also obtain the channel attention map $Z_c \in \mathbb{R}^{C\times 1\times 1}$ for the video frame by using the NCC operation along spatial dimension. Thus, the final spatial-channel attention map is calculated as follows:

\begin{equation}
\boldsymbol{Z} = sigmoid(\boldsymbol{Z}_s \odot \boldsymbol{Z}_c)
\end{equation}
The spatial-channel attention map is multiplied with the input representation $\boldsymbol{f}_{t,y}$ of the video frame to activate the common salient regions among video frames and focus on the common important channels. The output salient features of pedestrian are passed on to the spatial-temporal interaction module.

\subsection{Spatial-Temporal Interaction Module} 

\begin{figure}[!t]
	\begin{center}
		\includegraphics[width=1.0 \linewidth]{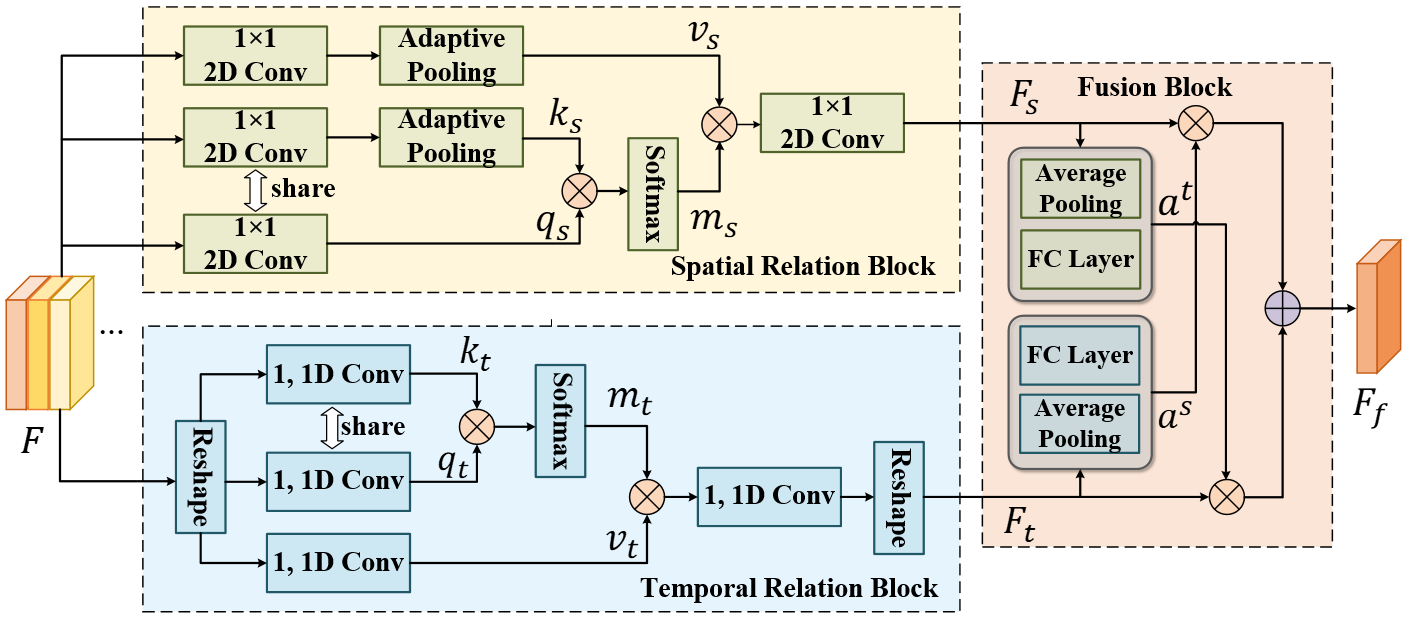}
	\end{center}
	\caption{Detailed structure of the spatial-temporal interaction module.}
	\label{img4}
\end{figure}

The spatial-temporal interaction module is designed for exploring the spatial-temporal long-range context interdependency on the common salient regions among video frames to enhance the capacity of the learned representation. As illustrated in Figure~\ref{img4}, the module consists of a spatial relation block, a temporal relation block and a fusion block. The spatial relation block and the temporal relation block are used to generate the primitive spatial and temporal relation features according to the learned semantic relation maps. Subsequently, an fusion block is utilized to learn the correlation between the primitive spatial and temporal relation features, and output a more effective pedestrian representation.

The spatial relation block consists of four convolution layers and two adaptive pooling layers. It discovers the spatial long-range interdependency between positions in the salient feature maps of video frames to obtain the primitive spatial relation features by using the non-local operation \cite{wang2018non}. The spatial relation block takes the output salient feature map from the co-saliency learning module as input, which is denoted by $\boldsymbol{F} \in \mathbb{R}^{T\times C\times H\times W}$. The input feature map is fed into three convolution layers and two adaptive average pooling layers to obtain feature maps $\boldsymbol{q}_s$, $\boldsymbol{k}_s$ and $\boldsymbol{v}_s$, respectively. The kernel sizes of the three convolution layers are $1 \times 1 \times C_1$. Moreover, the parameters of the two convolution layers for obtaining $\boldsymbol{q}_s$ and $\boldsymbol{k}_s$ are shared. Two adaptive pooling layers are used to reduce the computation cost. The output size of them are both $H_1 \times W_1$. Afterwards, the spatial relation map containing the spatial semantic relation between positions in the feature map is obtained by:

\begin{equation}
\boldsymbol{m_s} = Softmax(\boldsymbol{k}_s^T\boldsymbol{q}_s)
\end{equation}
Finally, the primitive spatial relation feature $\boldsymbol{F}_s$ is calculated as follows:
\begin{equation}
\boldsymbol{F}_s = \delta(\boldsymbol{G}_s\ast(\boldsymbol{v}_s\boldsymbol{m}_s))
\end{equation}
where $\delta$ refers to rectified linear unit operation, $\boldsymbol{G}_s$ and $\ast$ refer to the kernel of the last
convolution layer and convolution operation, respectively. The temporal relation block consists of four convolution layers. It is designed to exploit the temporal long-range interdependency between the same positions in the salient feature maps across video frames to obtain the primitive temporal relation feature. The input feature map  $\boldsymbol{F} \in \mathbb{R}^{T\times C\times H\times W}$ is first reshaped into  $\boldsymbol{F} \in \mathbb{R}^{HW \times C\times T}$. Then, $\boldsymbol{F}$ is fed into three convolution layers to generate feature maps $\boldsymbol{q}_t$, $\boldsymbol{k}_t$ and $\boldsymbol{v}_t$, respectively. The kernel sizes of the three convolution layers are  $1 \times 1 \times C_1$. Similar to Eq~(4),(5), we can obtain the temporal relation map $\boldsymbol{m}_t$ and the primitive temporal relation feature $\boldsymbol{F}_t$.

The fusion block consists of two fully connected layers and two global average pooling layers. It is used to learn the correlation between the spatial relation feature and temporal relation feature. The two global average pooling layers and two fully connected layers followed with the sigmoid functions are utilized to obtain the weight factors $\boldsymbol{a}_s$, $\boldsymbol{a}_t$ for the spatial and temporal relation features. The formulation is defined as follows:

\begin{equation}
\boldsymbol{a}_s = sigmoid(\boldsymbol{W}_sPooling(\boldsymbol{F}_t))
\end{equation}
The aggregated feature $\boldsymbol{F}_f$ is calculated by rescaling $\boldsymbol{F}_s$ and $\boldsymbol{F}_t$ with their corresponding weight factors:
\begin{equation}
\boldsymbol{F}_f^c = H_{scale}(\boldsymbol{F}^c_s, a_s^c) + H_{scale}(\boldsymbol{F}^c_t, a_t^c) 
\end{equation}
where $\boldsymbol{F}_f = [\boldsymbol{F}_f^1, \boldsymbol{F}_f^2,...,\boldsymbol{F}_f^C]$, $H_{scale}$ refers to channel-wise multiplication. The generated feature $\boldsymbol{F}_f$ is added with the feature of the current residual layer, and is finally fed into the next residual layer for further feature learning.

\subsection{Loss Function and Optimization} 

Identification loss and triplet loss are the commonly-used losses in the task of person re-identification, which have advantages in terms of simplicity and effectiveness. Accordingly, we adopt triplet loss with hard mining strategy \cite{hermans2017defense} and identification loss with label smoothing regularization \cite{szegedy2016rethinking} to optimize the proposed CSTNet. The total loss for the proposed CSTNet is the sum of triplet loss and identification loss. We randomly sample $P$ identities and $K$ video clips for each identity (each clip contains $T$ frames) to form a batch data. For the triplet loss, each sample, the corresponding hardest positive sample (the same pedestrian with large variance) and the hardest negative sample (similarly-looking but different pedestrians) in the batch are selected to form a triplet for computing the triplet loss. For the identification loss, the smoothing parameter $\epsilon$ is set to $0.1$ in our experiment.

\section{Experiments}

In this section, we conduct extensive experiments to evaluate the performance of the proposed CSTNet on two
video datasets and compare CSTNet with state-of-the-art methods. Moreover, we investigate the effectiveness of the proposed CSTNet and its components.

\begin{table}[!t]
	\begin{center}
		\newcommand{\tabincell}[2]{\begin{tabular}{@{}#1@{}}#2\end{tabular}}
		\resizebox{\columnwidth}{!}{
			\renewcommand{\arraystretch}{1.3}
			\begin{tabular}{c|c|c|c|c}
				\hline
				\textbf{Method}&\textbf{Rank-1}&\textbf{Rank-5}&\textbf{Rank-20}&\textbf{mAP}\\
				\hline
				MGCAM\cite{song2018mask}&77.2&-&-&71.2\\
				Triplet\cite{hermans2017defense}&79.8&91.4&-&67.7\\
				\hline
				JST-RNN \cite{zhou2017see}&70.6&90.0&97.6&50.7\\
				SpaAtn \cite{li2018diversity}&82.3&-&-&65.8\\
				STMP\cite{liu2019spatial}&84.4&93.2&96.3&72.7\\
				Snippet \cite{chen2018video}&86.3&94.7&98.2&76.1\\
				ADFA \cite{zhao2019attribute}&87.0&95.4&98.7&78.2\\
				GLTR \cite{li2019global}&87.0&95.8&98.2&78.5\\
				VRSTC \cite{hou2019vrstc}&88.5&96.5&-&82.3\\
				\hline 
				CSTNet &\textbf{90.2}&\textbf{96.8}&\textbf{98.7}&\textbf{83.9}\\
				\hline
		\end{tabular}}
	\end{center}
  \caption{Performance comparison to the state-of-the-art methods on the MARS dataset.}
	\label{table1}
\end{table}

\begin{table}[!t]
	\scriptsize
	\begin{center}
		\newcommand{\tabincell}[2]{\begin{tabular}{@{}#1@{}}#2\end{tabular}}
		\resizebox{\columnwidth}{!}{
			\renewcommand{\arraystretch}{1.2}
			\begin{tabular}{c|c|c|c}
				\hline
				\textbf{Method}&\textbf{Rank-1}&\textbf{Rank-5}&\textbf{Rank-20}\\
				\hline
				JST-RNN\cite{zhou2017see}&55.2&86.5&97.0\\
				RCN \cite{zhou2017see}&58.0&84.0&96.0\\
				ASPTN\cite{xu2017jointly}&62.0&86.0&98.0\\
				SpaAtn \cite{li2018diversity}&80.2&-&-\\
				VRSTC \cite{hou2019vrstc}&83.4&95.5&99.5\\
				STMP \cite{liu2019spatial}&84.3&96.8&99.5\\
				Snippet \cite{chen2018video}&85.4&96.7&99.5\\
				GLTR\cite{li2019global}&86.0&98.0&-\\
				ADFA \cite{zhao2019attribute}&86.3&97.4&\textbf{99.7}\\
				\hline 
				CSTNet &\textbf{87.8}&\textbf{98.5}&99.6\\
				\hline
		\end{tabular}}
	\end{center}
	\caption{Performance comparison to the state-of-the-art methods on the iLIDS-VID dataset.}
	\label{table2}
\end{table}

\begin{table}[!t]
	\scriptsize
	\begin{center}
		\newcommand{\tabincell}[2]{\begin{tabular}{@{}#1@{}}#2\end{tabular}}
		\resizebox{\columnwidth}{!}{
			\renewcommand{\arraystretch}{1.2}
			\begin{tabular}{c|c|c|c|c}
				\hline
				\textbf{Model}&\textbf{Rank-1}&\textbf{Rank-5}&\textbf{Rank-20}&\textbf{mAP}\\
				\hline
				Base&86.0&94.5&97.2&78.2\\
				\hline
				Base+ChanAtt&86.9&95.2&97.5&79.0\\
				Base+SpatAtt&87.8&95.8&97.9&81.5\\
				Base+CSL&88.2&96.1&98.0&81.8\\
				\hline
				Base+TemBlock&87.6&95.4&97.8&78.8\\
				Base+SpaBlock&88.1&95.8&98.0&80.4\\
				Base+STI&89.0&96.5&98.4&82.8\\
				\hline 
				CSTNet &90.2&96.8&98.7&83.9\\
				\hline
		\end{tabular}}
	\end{center}
	\caption{Evaluation of the effectiveness of each component within
	CSTNet on the MARS dataset.}
	\label{table3}
\end{table}


\subsection{Experimental Settings}

\paragraph{Datasets.} We evaluate the proposed CSTNet on two commonly used video-based person re-identification datasets: MARS and iLIDS-VID. The MARS dataset contains 1,261 identities and a total of 20,715 video sequences captured by 6 cameras. Each identity is captured by at least 2 cameras and has 13.2 video sequences on average. There are 3,248 distractor sequences in the dataset due to the failure of detection or tracking. The dataset is fixedly split into 625 identities for training and 636 identities for testing. The iLIDS-VID dataset consists of 600 video sequences of 300 pedestrians. For each pedestrian, there are two video sequences captured from two cameras views at an airport arrival hall. Each sequence has a variable length between 23 and 192 frames with an average length of 73. It is very challenging due to similar clothing among different pedestrians, blur, occlusions and viewpoint variations \textit{etc}. Following the implementation in the previous work \cite{wang2014person}, the dataset is randomly divided into a training set of 150 pedestrians and a testing set of 150 pedestrians.

\paragraph{Evaluation Metrics.} Cumulative Matching Characteristic (CMC) is extensively adopted for quantitative evaluation of person re-identification methods. The rank-$k$ recognition rate in the CMC curve indicates the probability that a query identity appears in the top-$k$ position. The other evaluation metric is the mean average precision (mAP), which is used to evaluate methods in multi-shot re-identification settings.

\paragraph{Implementation Details.} The implementation of the proposed method is based on the Pytorch framework with two Titan RTX GPUs. The input video frames are re-scale to the size of $3\times256\times 128$ and normalised with 1.0/256. The training set is enlarged by data augmentation strategies including random horizontal flipping and random erasing probability of 0.3. The parameters of $C_L$, $C_1$, $H_1$ and $W_1$ are set to 256, 128, 16 and 8 respectively. Each min-batch contains 16 identities and 4 video clips for each identity. Each video clip samples 8 video frames. The Adam optimizer is adopted with the learning rate $lr$ of $3e^{-4}$, the weight decay of $5e^{-4}$ and the Nesterov momentum of 0.9. The model is trained for 600 epochs in total. The learning rate $lr$ is multiplied by 0.1 after every 200 epochs.

\subsection{Comparison to State-of-the-Arts}

\paragraph{MARS:} Table~\ref{table1} shows the performance comparison of the proposed CSTNet against 9 state-of-the-art methods in terms of CMC accuracy and mAP score, including MGCAM \cite{song2018mask}, Triplet \cite{hermans2017defense}, JST-RNN \cite{zhou2017see}, SpaAtn \cite{li2018diversity},  STMP \cite{liu2019spatial}, Snippet \cite{chen2018video}, ADFA \cite{zhao2019attribute}, GLTR \cite{li2019global} and VRSTC \cite{hou2019vrstc}. The first two approaches are image-based person re-identification methods, and the others are video-based person re-identification methods. From the results, we can see that the video-based methods obtain much better performance in terms of both recognition rate and mAP score as compared to the image-based methods. The proposed CSTNet achieves 90.2\% rank-1 recognition rate and 83.9\% mAP score. We can see that CSTNet surpasses the existing methods, improving the 2nd best compared method VRSTC by 1.7\% rank-1 recognition rate and 1.6\% mAP score, respectively. The comparison indicates the effectiveness of the proposed CSTNet for modeling the spatial-temporal long-range interdependency on the salient feature maps of video frames to learn discriminative pedestrian representation.

\paragraph{iLIDS-VID:} We compare the proposed CSTNet against 9 state-of-the-art methods including: JST-RNN \cite{zhou2017see}, RCN \cite{zhou2017see}, ASPTN \cite{xu2017jointly}, SpaAtn \cite{li2018diversity}, VRSTC \cite{hou2019vrstc}, STMP \cite{liu2019spatial}, Snippet \cite{chen2018video}, GLTR \cite{li2019global} and ADFA \cite{zhao2019attribute}, which are designed for video-based person re-identification. From Table \ref{table2}, we can observe that the proposed CSTNet obtains the best 87.8\% rank-1 recognition rate and 98.5\% rank-5 recognition rate, outperforming all the existing methods at all ranks except for rank-20 recognition rate. The comparison indicates the effectiveness of the proposed CSTNet on the relatively small video person re-identification dataset.

\subsection{Ablation Studies} 

\paragraph{The impact of the CSL and STI modules.} Table~\ref{table3} summarizes the ablation results of the proposed CSTNet. Base, Base+CSL, Base+STI refer to the backbone network, the backbone network with the co-saliency learning module and the backbone network with the spatial-temporal interaction module, respectively. These three models obtain 86.0\%, 88.2\% and 89.0\% rank-1 recognition rate, as well as 78.2\%, 81.8\%, 82.8\% mAP score, respectively. The performance improvement of Base+CSL and Base+STI over Base, indicates that the two designed modules are able to generate more effective representations from raw pedestrian videos by focusing on the common salient regions and utilizing the spatial-temporal long-range interdependency. In addition, the performance of Base+CSL and Base+STI are inferior to CSTNet, shows that the effectiveness of CSTNet for joint exploration of the complementary two modules.

\paragraph{The impact of each component in the CSL module.} We conduct the experiments to verify the influence of each component in the co-saliency learning module. Base+ChanAtt and Base+SpatAtt refer to the backbone network with using the channel attention map and spatial attention map of the module, respectively. As shown in Table~\ref{table3}, Base+ChanAtt and Base+SpatAtt improve the Base by 0.9\%, 1.8\% rank-1 recognition rate, respectively, which shows the co-saliency learning module can learn precise spatial attention for the common salient regions among video frames and precise channel attention for the common important channels of feature maps among video frames. Moreover, the performance improvement of Base+CSL over Base+ChanAtt and Base+SpatAtt, shows that the effectiveness of the module for jointly learning the spatial-channel attention map.

\paragraph{The impact of each component in the STI module.} We also conduct the experiments to verify the effectiveness of each component in the spatial-temporal interaction module. Base+TemBlock and Base+SpaBlock refer to the backbone network with the temporal relation block and the spatial relation block of the module, respectively. From Table~\ref{table3}, it can seen that Base+TemBlock and Base+SpaBlock improve Base by 1.6\%, 2.1\% rank-1 recognition rate, respectively, indicates that the two blocks are able to learn more discriminative features by exploring the temporal and spatial long-range interdependency on the common salient regions among video frames. Moreover, the performance of the two blocks is inferior to Base+STI, shows that the effectiveness of the module for aggregating the spatial relation feature and temporal relation feature based on their correlation.

\section{Conclusion} 
In this work, we propose a novel Co-Saliency Spatio-Temporal Interaction Network (CSTNet) to learn discriminative pedestrian representation for person re-identification in videos. The co-saliency learning module within the CSTNet captures the common salient foreground regions across video frames by utilizing the normalized cross correlation for extracting salient feature maps and suppressing background interference. Moreover, the spatial-temporal interaction module exploits the spatial and temporal long-range context interdependencies on such common salient regions and spatial-temporal information correlation to enhance the effectiveness of pedestrian representation. Extensive experiments on two challenging benchmarks have shown that the proposed CSTNet achieves significant performance improvement over a wide range of state-of-the-art methods.

\section*{Acknowledgments}
This work was supported by the National Key R\&D Program of China under Grant 2017YFB1300201, the National Natural Science Foundation of China (NSFC) under Grants U19B2038 and 61620106009 as well as the Fundamental Research Funds for the Central Universities under Grant WK2100100030. 


\bibliographystyle{named}
\bibliography{ijcai20}

\end{document}